\documentclass[11pt]{article}

\usepackage[final]{acl}

\usepackage{times}
\usepackage{latexsym}
\usepackage[T1]{fontenc}
\usepackage[utf8]{inputenc}
\usepackage{microtype}
\usepackage{inconsolata}
\usepackage{graphicx}
\usepackage{amssymb}
\usepackage{amsmath}
\usepackage{booktabs}
\usepackage{multirow}

\title{MIRTH: Mutual-Information Reasoning with Temporal Hubs for Vision-Language-Action Agents}

\author{
 \textbf{Hao Sun\textsuperscript{1}},
 \textbf{Yu Song\textsuperscript{1}},
 \textbf{Shiyu Teng\textsuperscript{1}},
 \textbf{Ziwei Niu\textsuperscript{2}},
 and 
 \textbf{Yen-Wei Chen\textsuperscript{1}}
\\
 \textsuperscript{1}College of Information Science and Engineering, Ritsumeikan University, Japan,\\
 \textsuperscript{2}College of Computer Science and Technology, Zhejiang University, China,
\\
 \small{
   \textbf{Correspondence:} \href{mailto:sunhaoxx@fc.ritsumei.ac.jp}{sunhaoxx@fc.ritsumei.ac.jp} / \href{mailto:sunhaoxx@zju.edu.cn}{sunhaoxx@zju.edu.cn} (Hao Sun) and \href{mailto:chen@is.ritsumei.ac.jp}{chen@is.ritsumei.ac.jp (Yen-Wei Chen)}
 }
}

\begin{document}
\maketitle
\begin{abstract}
VLA models have emerged as a powerful paradigm for transferring semantic knowledge from web-scale data to physical robotic control. However, current single-frame architectures suffer from intrinsic limitations: temporal myopia that discards historical dynamics, reasoning gaps between high-level instructions and low-level motor commands, and inference inefficiency due to autoregressive scalar decoding. In this work, we propose MIRTH, a unified framework designed to address these challenges. MIRTH augments a pretrained VLA backbone with three key innovations: (1) dual-scale temporal memory hubs that compress long-term scene evolution and short-term motion trends into compact embeddings; (2) latent reasoning tokens optimized via a mutual-information objective carving out a semantic plan space to align multimodal context with action trajectories; and (3) a parallel action decoding scheme that replaces autoregressive generation with vector-wise prediction to maximize control throughput. Extensive evaluations on the LIBERO simulation benchmark and a real-world LeRobot platform demonstrate that MIRTH achieves state-of-the-art performance and exhibiting emergent error recovery capabilities. The codes and collected datasets are released at \textit{http://github.com/kiva12138/mirth}.
\end{abstract}

\section{Introduction}
Robotic agents capable of following open-ended natural language instructions, perceiving complex visual scenes, and executing fine-grained motor skills hold the promise of transforming human-machine interaction in unstructured environments. Recent Vision–Language–Action (VLA) models, built upon the success of Large Vision–Language Models (VLMs) and massive cross-embodiment datasets \citep{o2024open}, have made significant strides in this direction. By treating robot actions as essentially another \textit{language} and leveraging web-scale pretraining, models such as RT-2 \citep{zitkovich2023rt}, PaLM-E \citep{driess2023palme}, and OpenVLA \citep{kim2024openvla} have demonstrated remarkable capabilities, effectively transferring semantic knowledge from Internet data to physical control. However, despite these advances, current open-source VLA architectures face intrinsic structural limitations that hinder their applicability to long-horizon and dynamic tasks.

\begin{figure}[t]
    \centering
    \includegraphics[width=\linewidth]{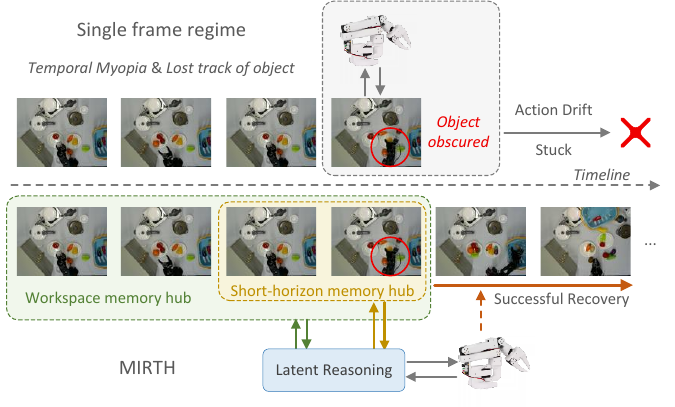}
    \caption{\textbf{Overcoming temporal myopia with MIRTH.} Standard single-frame VLA models (e.g., OpenVLA) suffer from temporal myopia. When objects get obscured during manipulation, the agent loses track of the object state, leading to execution failure. MIRTH introduces two memory hubs to actively track long-term scene layout and short-term dynamics. Coupled with latent reasoning tokens, MIRTH successfully maintains the object's position in memory despite occlusion, enabling robust recovery and successful completion.}
    \label{fig:teaser}
\end{figure}

We identify three critical challenges that limit the efficacy of current VLA paradigms. First, existing opensource models suffer from \textbf{temporal myopia}. Most state-of-the-art VLAs operate in a single-frame regime, decoding actions conditioned solely on the immediate observation and instruction (shown in Figure~\ref{fig:teaser}). This Markovian assumption, common in architectures like RT-1 \citep{brohan2022rt} and Q-Transformer \citep{chebotar2023q}, discards rich temporal cues that are essential for robust decision-making, such as motion trends and object permanence during occlusions. While recent approaches like ACT \citep{zhao2023learning} incorporate temporal chunking, they often lack the generalizable semantic understanding of VLMs. Second, there is a \textbf{reasoning gap} in connecting high-level linguistic goals to low-level motor commands. Prior works typically map visual-language inputs directly to actions or utilize discrete action tokens supervised by hand-crafted vocabularies. The former lacks interpretability and internal planning \citep{wei2022chain}, while the latter suffers from the \textit{many-to-one} problem, where diverse linguistic descriptions map to identical physical motions \citep{lee2024behavior}. Third, the autoregressive generation of continuous actions imposes a severe \textbf{efficiency bottleneck}. Standard strategies quantize continuous action dimensions into discrete tokens, forcing the model to emit a long sequence of tokens for a single maneuver. This results in high decoding latency, rendering real-time, high-frequency control computationally prohibitive compared to diffusion-based policies \citep{mees2024octo}.

To address these challenges within a unified framework, we propose \textbf{MIRTH} (\textbf{M}utual-\textbf{I}nformation \textbf{R}easoning with \textbf{T}emporal \textbf{H}ubs). MIRTH augments a pretrained VLA backbone with a novel architecture designed to bridge the gap between multimodal context, latent reasoning, and efficient execution. Specifically, our framework introduces three key innovations. (1) We introduce dual-scale \textbf{temporal memory hubs} that compress long-term scene layouts (via workspace memory) and short-term motion dynamics into fixed-length prompts. This enables the models to condition on arbitrary histories to overcome temporal myopia without inflating the context window. (2) We also introduce a set of \textbf{latent reasoning tokens} optimized via a mutual-information objective. These tokens carve out a semantic plan space to align visual observations with action trajectories without relying on expensive and ambiguous text supervision. (3) A \textbf{parallel action decoding} scheme is utilized, which replaces scalar-wise autoregression with vector-wise prediction, significantly reducing decoding latency to ensure real-time execution efficiency. By explicitly structuring memory and reasoning within the token space, MIRTH maintains the semantic richness of VLMs while enabling precise, history-aware control for physical agents.

Finally, instead of utilizing expensive and closesource robot platforms, we evaluate MIRTH on opensource simulation and embodied platforms. The challenging LIBERO simulation benchmark suite \citep{liu2023libero} and a open-source LeRobot platform\footnote{https://github.com/huggingface/lerobot} are employed in our experiments. We cover tasks necessitating from long-horizon dependency to multi-step reasoning. MIRTH consistently outperforms strong single-frame and naive multi-frame baselines, achieving near-perfect success rates on LIBERO benchmarks and demonstrating superior robustness in real-world deployments. Ablation studies confirm that structuring memory and enforcing latent reasoning are crucial for these gains. Qualitative analysis further reveals that MIRTH's reasoning tokens emerge as meaningful semantic clusters, enabling the agent to re-plan dynamically upon failure. 
To facilitate further research, we will release all codes and datasets upon publication.

\begin{figure*}
    \centering
    \begin{minipage}{1.0\linewidth}
        \centering
        \includegraphics[width=1.0\textwidth]{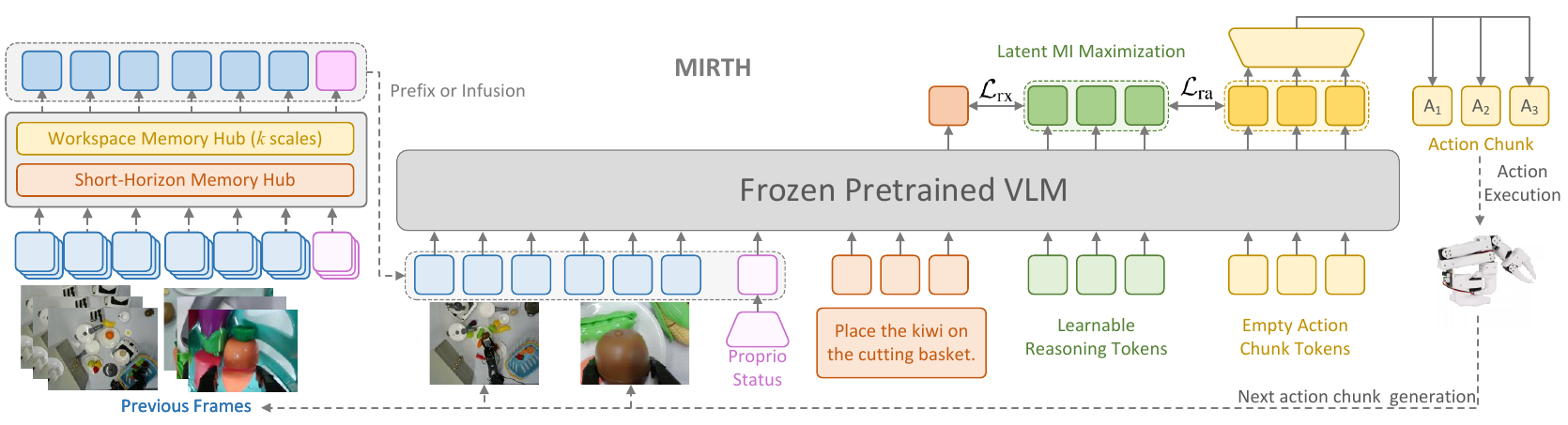}
    \end{minipage}
  \caption{\textbf{The overall pipeline of MIRTH}. To effectively integrate historical context, we propose temporal memory hubs, comprising a long-term workspace hub and a short-horizon hub. The fused historical features are integrated into the current frame's representation via either token prefixing or patch infusion. Crucially, we introduce a set of Latent Reasoning Tokens optimized to maximize the mutual information between the environmental context and action embeddings, serving as a compact planning bridge. Finally, the full sequence of action tokens is generated via Parallel Action Decoding for efficient robot execution.}
  \label{fig:pipeline}
\end{figure*}

\section{Related Works}
\subsection{Vision-Language-Action Models}
Recent VLA architectures have successfully repurposed web-scale VLMs for robotic control by treating actions as tokens in an autoregressive sequence. Models like RT-2 \citep{zitkovich2023rt} and PaLM-E \citep{driess2023palme} demonstrated that LLMs can directly output robot commands when fine-tuned on multimodal data. More recently, open-source efforts such as OpenVLA \citep{kim2024openvla} and RoboFlamingo \citep{li2023roboflamingo} have democratized access to these capabilities, leveraging visual encoders like DINOv2 \citep{oquab2024dinov2}, SigLIP \citep{zhai2023sigmoid}, and LLaVA \citep{liu2024llava}. However, these models typically employ a single-frame paradigm, mapping immediate visual observations directly to action tokens. While effective for short-horizon tasks, this design is inherently myopic. Naively extending context windows to capture temporal dynamics leads to prohibitive computational costs \citep{ashton2025helios, o2024open}, rendering such models brittle in scenarios requiring long-term object tracking or error recovery.

An alternative approach to integrating historical context is the adoption of standard Video Transformers~\cite{Pizarro2026-jd,patraucean2026trecvitrecurrentvideotransformer}. However, because these models rely on dense spatio-temporal attention across multiple frames, they inherently suffer from quadratic computational scaling. This complexity results in a massive memory footprint and severe inference latency, violating the strict real-time control requirements essential for robotics. In contrast, MIRTH explicitly circumvents this bottleneck. By compressing long-term and short-term history into fixed-length prompts via dual-scale temporal memory hubs, our design prioritizes inference efficiency. This allows MIRTH to achieve a high control throughput, making it highly suitable for deployment on edge hardware without sacrificing temporal awareness.

\subsection{Intermediate Reasoning in Robotics}
To bridge the semantic gap between high-level instructions and low-level control, prior works often enforce explicit reasoning steps. Methods like SayCan \citep{ahn2022can} and Code as Policies \citep{liang2022code} decompose tasks into textual subgoals or generated codes. Other methods such as VoxPoser \citep{huang2023voxposer} and RT-Trajectory \citep{gu2023rt1} supervise the model with discrete \textit{action language} tokens or value maps. However, these approaches suffer from the high cost of collecting dense textual annotations and the ambiguity of natural language, where many descriptions map to the same trajectory. In contrast, MIRTH diverges from this paradigm by eschewing explicit text supervision for intermediate steps. Instead, we employ a mutual-information objective to learn latent reasoning tokens that naturally align visual context with action intentions, similar to unsupervised skill discovery \citep{shafiullah2022behavior}, but within a VLA framework without requiring expensive human annotation.

\subsection{Action Decoding Efficiency}Standard VLAs ground continuous actions via scalar-wise quantization, where each degree of freedom corresponds to a discrete vocabulary token \citep{zitkovich2023rt,brohan2022rt}. While this unifies the input-output space, it creates a severe bottleneck: generating a single pose requires an autoregressive chain of tokens for every action dimension, limiting control frequency. Although recent diffusion-based policies \citep{chi2025diffusion} offer alternatives, they often sacrifice the benefits of a unified transformer architecture. MIRTH resolves this tension by maintaining a token-based interface but shifting to a parallel, vector-wise decoding scheme. This design eliminates the autoregressive overhead for action dimensions, enabling high-throughput control suitable for real-time deployment.

\section{MIRTH}
In this section, we present the details of our proposed MIRTH, including multi-frame integration, mutual-information reasoning, and optimized action token decoding. The overview of MIRTH is shown in Figure~\ref{fig:pipeline}. \footnote{The accepted version of the paper contains errors regarding symbols and repetitions; please refer to this latest version uploaded to arXiv.}

\subsection{Temporal Memory Hubs}
Building on the pretrained single-frame VLA backbone, we explicitly exploit motion trends and object dynamics by introducing two temporal memory hubs that integrate information from multiple past observations. For clarity, we describe the construction on visual features; the same design is applied to proprioception in an analogous way. At each timestep $t$, we denote the patch-level embedding (the patch-number dimension is omitted for convenience) of the current multi-camera frame as $\mathbf{V}_t \in \mathbb{R}^{N \times D}$, obtained by DINOv2+SigLIP encoder.

To summarize long-range history, we maintain a \emph{workspace (long-term) memory hub} composed of $K$ exponential moving averages with different decay rates $\{\beta_k\}_{k=1}^K$, where each scale $k$ stores a memory map $\mathbf{M}_{t,k} \in \mathbb{R}^{N \times D}$:
\begin{equation}
    \mathbf{M}_{t,k} = (1-\beta_k)\,\mathbf{M}_{t-1,k} + \beta_k \mathbf{V}_t.
\end{equation}
In parallel, we track first- and second-order temporal statistics of per-patch changes
\begin{equation}
\begin{aligned}
    \Delta_t &= \mathbf{V}_t - \mathbf{V}_{t-1}\in \mathbb{R}^{N \times D}, \\
    \mathbf{\mu}_t &= (1-\gamma_\mu)\,\mathbf{\mu}_{t-1} + \gamma_\mu\,\Delta_t \in \mathbb{R}^{N \times D},\\
    \mathbf{\sigma}_t^2 &= (1-\lambda_\sigma)\,\mathbf{\sigma}^2_{t-1} + \lambda\sigma\,(\Delta_t \odot \Delta_t) \in \mathbb{R}^{N \times D},
\end{aligned}
\end{equation}
where $\mathbf{\mu}_t$ and $\mathbf{\sigma}^2_t$ capture local motion velocity and variability, and $\odot$ denotes element-wise multiplication. At timestep $t$, we form a descriptor and use a small MLP to predict per-patch mixture weights over the $K$ time scales:
\begin{equation}
\begin{aligned}
    \boldsymbol{\alpha}_t &= \mathrm{softmax}\!\big(\mathbf{W}_a [\mathbf{V}_t;\,\mathbf{\mu}_t;\,\mathbf{\sigma}_t]\big) \in \mathbb{R}^{N \times K},
\end{aligned}
\end{equation}
where $\mathbf{W}_a \in \mathbb{R}^{K \times 3D}$. The workspace memory is then a mixture of multi-scale memories plus projected motion statistics:
\begin{equation}
    \mathbf{M}_t^{\text{work}} = \sum_{k=1}^K \alpha_{t,k} \odot \mathbf{M}_{k,t} 
    \;+\; \mathbf{W}_\mu \mathbf{\mu}_t \;+\; \mathbf{W}_\sigma \mathbf{\sigma}_t,
\end{equation}
yielding a long-term feature bank $ \mathbb{R}^{N \times D}$ that can adaptively emphasize either slowly varying context (e.g., scene layout) or fast dynamics (e.g., moving directions), without storing all past frames explicitly \citep{wu2019long,fan2021multiscale,weston2014memory}.

Complementary to this, we introduce a \emph{short-horizon (short-term) memory hub} that focuses on the most recent $w$ frames. We maintain a fixed-length queue
\begin{equation}
    \mathbf{Q}_t = \big[\mathbf{V}_{t-w+1},\dots,\mathbf{V}_t\big] \in \mathbb{R}^{w \times N \times D},
\end{equation}
and compute a temporal attention distribution. A query is derived from the current frame, and keys / values from the queue:
\begin{equation}
\begin{aligned}
    \mathbf{q}_t &= \mathbf{W}_q \mathbf{V}_t \in \mathbb{R}^{N \times D},\\
    \mathbf{K}_t^{\text{short}} &= \mathbf{W}_k \mathrm{LN}(\mathbf{Q}_t) \in \mathbb{R}^{w \times N \times D},\\
    \mathbf{V}_t^{\text{short}} &= \mathbf{W}_v \mathrm{LN}(\mathbf{Q}_t) \in \mathbb{R}^{w \times N \times D}.
\end{aligned}
\end{equation}
For each temporal index $j \in \{0,\dots,w-1\}$ (from oldest to most recent), we compute attention logits with a temperature $\tau_r$ and a recency bias $\gamma_r$:
\begin{equation}
\begin{aligned}
    \boldsymbol{\pi}_t &= \mathrm{softmax}_j(\frac{1}{\tau_r}\,\big\langle \mathbf{q}_t,\mathbf{K}_{t,j}^{\text{short}} \big\rangle + \gamma_r\, j) \in \mathbb{R}^{w \times N},
\end{aligned}
\end{equation}
where $\langle\cdot,\cdot\rangle$ denotes the dot product. The short-horizon memory is the attention-weighted sum of the value embeddings:
\begin{equation}
    \mathbf{M}_t^{\text{short}} = \sum_{j=0}^{w-1} \pi_{t,j} \odot \mathbf{V}_{t,j}^{\text{short}} \in \mathbb{R}^{N \times D}.
\end{equation}

Finally, we fuse the two hubs using a per-patch sigmoid gate $g_t$ computed from $[\mathbf{V}_t;\mathbf{M}_t^{\text{work}};\mathbf{S}_t^{\text{short}}]$:
\begin{equation}
\begin{aligned}
    g_t &= \sigma\big(\mathbf{W}_g [\mathbf{V}_t;\mathbf{M}_t^{\text{work}};\mathbf{M}_t^{\text{short}}]\big),\\
    \mathbf{M}_t^{\text{fused}} &= g_t \odot \mathbf{M}_t^{\text{work}} + (1-g_t) \odot \mathbf{M}_t^{\text{short}}.
\end{aligned}
\end{equation}
We then explore two integration strategies for the memory token map $\mathbf{M}_t^{\text{fused}}$. In the \emph{prefix} variant, the memory is flattened and prepended as a compact multi-frame visual prefix:
\begin{equation}
    \tilde{\mathbf{Z}}_t^{\text{prefix}} = [\mathrm{Flatten}(\mathbf{M}_t^{\text{fused}});\;\mathbf{Z}_t],
\end{equation}
where $\mathbf{Z}_t$ denotes the original sequence of visual (and proprioceptive) tokens. In the \emph{infusion} variant, we modulate the original patch embeddings multiplicatively and additively via two linear projections:
\begin{equation}
    \tilde{\mathbf{Z}}_t^{infus} = \mathbf{Z}_t \odot \big(1 + \mathbf{W}_{\mathrm{mult}}\mathbf{M}_t^{\text{fused}}\big) + \mathbf{W}_{\mathrm{add}}\mathbf{M}_t^{\text{fused}}.
\end{equation}
Both strategies have complementary advantages, and detailed discussions are illusrated in Appendix~\ref{sec:prefixvsinfusion}. 

Intuitively, the workspace hub behaves like a slowly evolving working memory that accumulates scene-level evidence over long horizons, whereas the short-horizon hub captures fine-grained recent motions.
Importantly, the memory size and the number of tokens fed into the language model remain fixed regardless of the history length, enabling multi-frame integration without compromising the efficiency constraints of real-world robotic control.

\subsection{Latent Reasoning}\label{sec:reasoning}
To avoid high-cost annotation while still benefiting from reasoning, we introduces a set of \emph{latent reasoning tokens} and trains them under a mutual-information objective, rather than forcing them to match any particular hand-crafted textual descriptions. Concretely, at each timestep $t$ we construct the multimodal input sequence to the language backbone as:
\begin{equation}
    \mathbf{X}_t = [\,\tilde{\mathbf{Z}}_t;\,\mathbf{L}_t;\,\mathbf{T}^{\text{reas}}_t;\,\mathbf{T}^{\text{act}}_t],
\end{equation}
where $\tilde{\mathbf{Z}}_t$ denotes the fused multi-frame visual/proprioceptive tokens from memory hubs, $\mathbf{L}_t$ are the tokenized language instructions, $\mathbf{T}^{\text{reas}}_t \in \mathbb{R}^{m \times D}$ is a small set of $m$ learnable reasoning tokens, and $\mathbf{T}^{\text{act}}_t$ are optimized action tokens. The reasoning tokens are thus positioned between \textit{condition} (perceptual context + instructions) and \textit{effect} (actions), and are intended to form a compact latent bridge between them.

Let $\mathbf{H}_t \in \mathbb{R}^{L \times D}$ denote the final hidden states output by the language backbone for $\mathbf{X}_t$, with the total sequence length $L$. From $\mathbf{H}_t$ we extract three pooled representations: (1) a reasoning representation $\mathbf{r}_i \in \mathbb{R}^{D}$ for each sequence $i$ by averaging over the hidden states corresponding to reasoning tokens; (2) an action representation $\mathbf{a}_i \in \mathbb{R}^{D}$ by averaging over the action-token positions; and (3) a context representation $\mathbf{x}_i \in \mathbb{R}^{D}$ from the hidden state immediately preceding the first reasoning token, which summarizes the upstream multimodal context $[\tilde{\mathbf{Z}}_t;\mathbf{L}_t]$. We project each of these into a shared contrastive space: $\mathbf{z}^R_i$, $\mathbf{z}^A_i$, $\mathbf{z}^X_i$; and interpret $\mathbf{z}^R_i$ as a latent reasoning code that should be maximally informative about both the current context and the corresponding action trajectory.

We formalize this intuition with a contrastive objective based on the InfoNCE loss \citep{oord2018representation}, which provides a lower bound on mutual information. Given a minibatch of $B$ sequences, we first encourage the reasoning representation to be predictive of the paired action representation by minimizing
\begin{equation}
    \mathcal{L}_{\text{ra}} = -\frac{1}{B} \sum_{i=1}^{B} 
    \log \frac{\exp\big(s(\mathbf{z}^R_i,\mathbf{z}^A_i)/\tau_{\text{ra}}\big)}
    {\sum_{j=1}^{B} \exp\big(s(\mathbf{z}^R_i,\mathbf{z}^A_j)/\tau_{\text{ra}}\big)},
\end{equation}
where $s(\cdot,\cdot)$ is a similarity function and $\tau_{\text{RA}}$ is a temperature hyperparameter. Symmetrically, we encourage the reasoning representation to be predictive of the multimodal context by minimizing
\begin{equation}
    \mathcal{L}_{\text{rx}} = -\frac{1}{B} \sum_{i=1}^{B} 
    \log \frac{\exp\big(s(\mathbf{z}^R_i,\mathbf{z}^X_i)/\tau_{\text{rx}}\big)}
    {\sum_{j=1}^{B} \exp\big(s(\mathbf{z}^R_i,\mathbf{z}^X_j)/\tau_{\text{rx}}\big)}.
\end{equation}
The overall mutual-information reasoning loss is a weighted sum
\begin{equation}
    \mathcal{L}_{\text{mi}} = \lambda_{\text{ra}}\,\mathcal{L}_{\text{ra}} + \lambda_{\text{rx}}\,\mathcal{L}_{\text{x}},
\end{equation}
which we add to the final loss for action prediction.

In this way, $\mathcal{L}_{\text{ra}}$ pulls reasoning tokens toward the space of action trajectories, while $\mathcal{L}_{\text{rx}}$ simultaneously anchors them in the multimodal context; together, they encourage the reasoning tokens to encode information that is jointly informative about \textit{what the world looks like} and \textit{what actions will be taken} without committing to any particular textual rationalization. Compared to directly supervising explicit action-language descriptions, this latent reasoning scheme is label-efficient and robust to the many-to-one mapping between language and control. 

\subsection{Action Decoding, Training, and Inference}
To address the latency bottleneck of standard VLA models where autoregressively generating scalar-quantized tokens for every action dimension scales poorly with trajectory length, we introduce a parallel action decoding paradigm with two key design choices. First, instead of representing each degree of freedom as an individual token, we allocate one action token per full action vector. Concretely, the action at timestep $t$ with $N_F$ degrees of freedom is represented by a single dedicated token position in the sequence, whose hidden state is later mapped to the continuous action space. For an action sequence with $N_A$ actions and $N_F$ degree of freedom, this reduces action-token sequence length by $1 / N_F$ compared to scalar-wise tokenization.

Second, given the hidden states at all action-token positions, we decode the entire action trajectory in parallel rather than in an autoregressive manner. Let $H^{\text{act}} \in \mathbb{R}^{N_A \times D}$ denote the final hidden states of action tokens. We apply a lightweight two-layer projection head which maps each action-token representation to the corresponding continuous action vector. This design separates reasoning from control decoding and enables efficient batched computation of all actions in a trajectory with a single forward pass. More detailed action decoding discussions are illusrated in Appendix~\ref{sec:action_decoding}

Training of the action decoder is performed with a standard regression objective on the continuous action space. Given the ground-truth action sequence $\hat{A} \in \mathbb{R}^{N_A \times N_F}$, we minimize the element-wise $\ell_1$ loss:
\begin{equation}
\mathcal{L}_{\text{l1}} = \frac{1}{N_A N_F} \sum_{i=1}^{N_A} \sum_{j=1}^{N_F} \bigl| A_{i,j} - \hat{A}_{i,j} \bigr|.
\end{equation}
The final training objective combines this regression loss with the mutual-information reasoning objective introduced in Section~\ref{sec:reasoning}:
\begin{equation}
    \mathcal{L} = \mathcal{L}_{\text{l1}} + \lambda_{\text{mi}} \mathcal{L}_{\text{MI}},
\end{equation}
where $\lambda_{\text{mi}}$ is a scalar weight balancing the strength of mutual-information regularization and action accuracy.

At inference time, MIRTH benefits from both the parallel action decoding and the temporally smoothed memory hubs. The EMA updates for visual and proprioceptive memory allow us to cache and incrementally update the multimodal embeddings across timesteps. As a result, for each new control step the model only needs to encode the current frame, update the memory hubs, and decode the actions parallelly in a single forward pass. This leads to markedly higher action-generation throughput compared to autoregressive action decoding, while preserving a unified token-based interface between perception, reasoning, and control.

\section{Results and Analysis}
We evaluate the effectiveness of MIRTH through a comprehensive set of experiments encompassing both simulated benchmarks and real-world manipulation tasks. Our evaluation is designed to answer the following core research questions:

\begin{itemize}
    \item \textbf{RQ1:} How does MIRTH compare against state-of-the-art open-source VLA baselines in terms of success rate and efficiency?
    \item \textbf{RQ2:} What is the contribution of each proposed component to the overall performance?
    \item \textbf{RQ3:} Do the temporal memory hubs effectively capture and retain motion dynamics and historical context compared to single-frame baselines?
    \item \textbf{RQ4:} Do the latent reasoning tokens enable the model to handle tasks requiring complex, multi-step reasoning and error recovery?
\end{itemize}

To rigorously assess MIRTH's performance in both standardized and unstructured environments, we conduct evaluations across two complementary domains: the widely-adopted LIBERO simulation benchmark and a physical LeRobot platform. Detailed experimental setups are illusrated in Appendix~\ref{sec:setup}.

\subsection{Evaluation Results}

\begin{table*}[ht]
\small
\centering
\caption{Evaluation results on LIBERO benchmarks. For each benckmark, the success rate are averaged over 500 eposides with different seeds. The results of other methods are from \cite{zhao2025cot} and \cite{kim2025fine}.}
\label{tab:libero_results}
\begin{tabular}{cccccc}
\toprule
 & Spatial & Object & Goal & Long & Average \\
\midrule
Diffusion Policy & 78.3 ± 1.1\% & 92.5 ± 0.7\% & 68.3 ± 1.2\% & 50.5 ± 1.3\% & 72.4\% \\
Octo & 78.9 ± 1.0\% & 85.7 ± 0.9\% & 84.6 ± 0.9\% & 51.1 ± 1.3\% & 75.1\% \\
OpenVLA & 84.7 ± 1.4\% & 88.4 ± 0.8\% & 79.2 ± 1.1\% & 53.7 ± 0.7\% & 76.5\% \\
OpenVLA-OFT & 97.6 ± 0.7\% & 98.4 ± 0.4\% & 97.9 ± 0.8\% & 94.5 ± 0.9\% & 97.1\% \\
MIRTH(ours) & \textbf{98.2 ± 0.6\%} & \textbf{100.0 ± 0.4\%} & \textbf{98.8 ± 0.5\%} & \textbf{95.3 ± 1.1\%} & \textbf{98.1} \%\\
\bottomrule
\end{tabular}
\end{table*}

\textbf{LIBERO Simulation Benchmark.} We present the quantitative comparison against SOTA open-source baselines, including Diffusion Policy \citep{chi2025diffusion}, Octo \citep{mees2024octo}, OpenVLA \citep{kim2024openvla}, and OpenVLA-OFT \cite{kim2025fine}, in Table~\ref{tab:libero_results}. As evidenced by the results, MIRTH demonstrates consistent superiority across all evaluation suites, achieving near-saturation performance on standard benchmarks. Specifically, on the spatial, goal, and object suites, MIRTH attains success rates exceeding 98\%, significantly outperforming the single-frame OpenVLA baseline. Most notably, in the challenging LIBERO-long suite, which necessitates maintaining context over extended horizons, MIRTH maintains a high success rate of 95\%. We attribute this robustness to the proposed temporal memory hubs: unlike the myopic single-frame baselines that fail to track state changes once objects are occluded or moved, MIRTH's EMA-based workspace hub effectively preserves the historical scene layout, while the latent reasoning tokens ensure that multi-step plans remain consistent over time. Some visualized qualitative comparisons are provided in Figure~\ref{fig:rollout_visualization}.

\begin{figure}
    \centering
    \begin{minipage}{1.0\linewidth}
        \centering
        \includegraphics[width=1.0\textwidth]{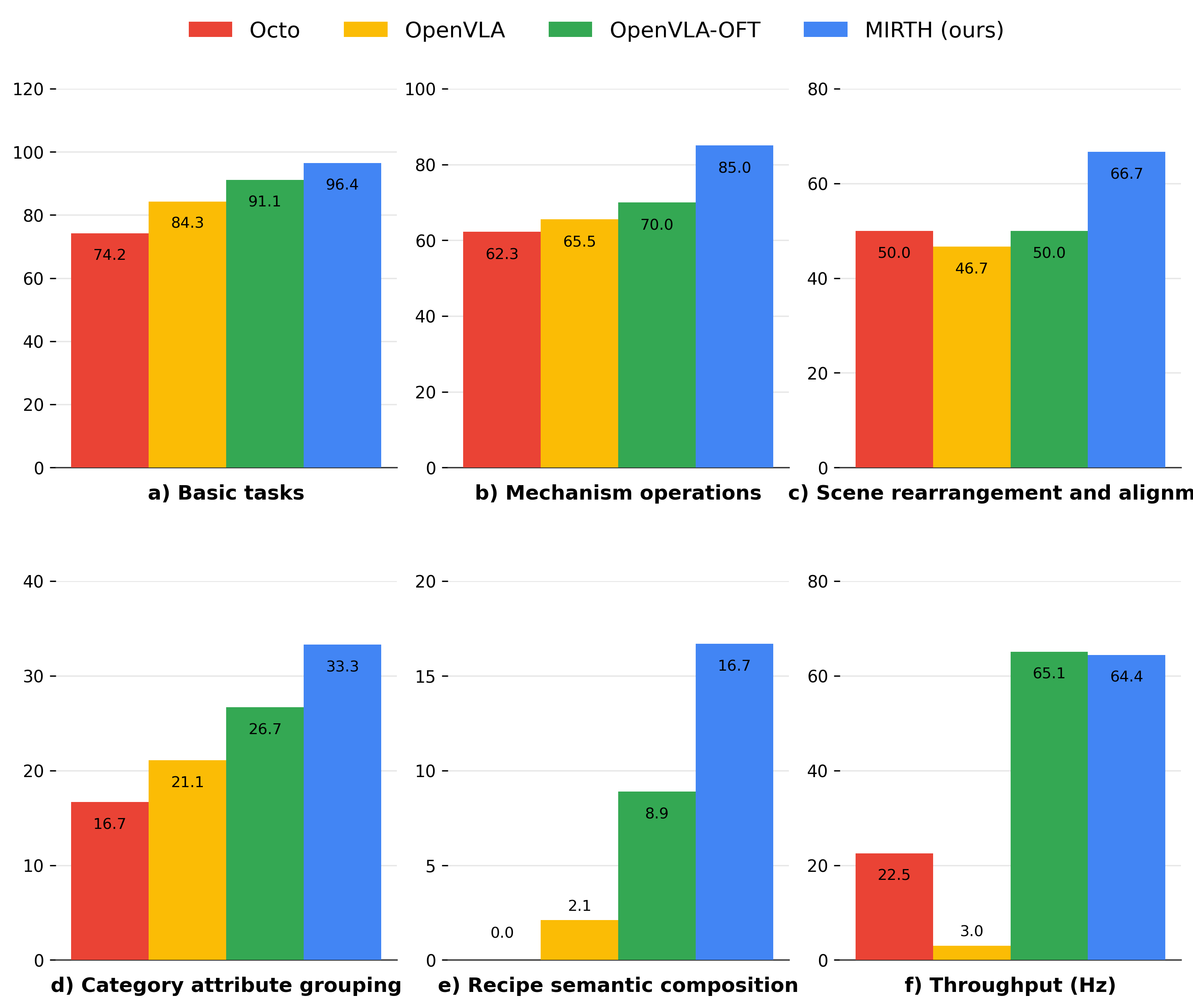}
    \end{minipage}
  \caption{\textbf{The comparison results on LeRobot across five task groups}. Results represent success rates averaged over 30 runs per task. MIRTH consistently achieves top-tier performance and throughput.}
  \label{fig:lerobot_resutls}
\end{figure}

\textbf{LeRobot Real-World Evaluation.} To assess scalability and reasoning capabilities in unstructured physical environments, we evaluate MIRTH on the LeRobot platform across five task groups of increasing complexity, ranging from atomic pick-and-place primitives to long-horizon manipulation requiring semantic reasoning. The methods are further fine-tuned on our collected trajectories. For detailed task descriptions, please refer to Appendix~\ref{sec:tasks}. The comparative results are summarized in Figure~\ref{fig:lerobot_resutls} and some visualizations are provided in Figure~\ref{fig:rollout_visualization}. MIRTH exhibits a widening performance gap compared to baselines as task difficulty increases. In complex reasoning scenarios, our method outperforms standard VLAs by a substantial margin where the agent must deduce the correct object based on abstract instructions. This improvement validates the effectiveness of our mutual-information reasoning objective, which successfully grounds abstract linguistic goals into precise physical actions without explicit step-by-step supervision. Furthermore, despite incorporating rich multi-frame context, MIRTH maintains a high inference throughput comparable to or exceeding lighter baselines (especially OpenVLA). This efficiency is directly driven by our parallel action decoding scheme and the compact design of the memory hubs.

\subsection{Ablation Studies} 
To validate the design choices of MIRTH, we conduct component-wise ablation studies on LIBERO-long. We systematically remove or modify key components including the workspace hub, short-term hub, reasoning kokens, and the mutual-information (MI). The results are shown in Table~\ref{tab:ablation}.

\begin{table}[t]
\small
\centering
\caption{Ablation Study of MIRTH on LIBERO-Long simulation benchmark. Success rate is employed as the main metric.}
\label{tab:ablation}
\begin{tabular}{lc}
\toprule
\textbf{Ablation} & \textbf{LIBERO-Long} \\
\midrule
\textbf{MIRTH (Full Model)} & \textbf{95.3\%} \\
\midrule
\multicolumn{1}{l}{\textit{Temporal Memory Hubs}} \\
\quad w/o workspace memory hub & 94.0\% \\
\quad w/o short-term memory hub & 94.4\% \\
\quad w/o both (single-frame baseline) & 93.2\%\\
\midrule
\multicolumn{1}{l}{\textit{Reasoning \& Alignment}} \\
\quad w/o MI loss (tokens w/o constraint) & 94.5\% \\
\quad w/o reasoning tokens & 93.9\% \\
\bottomrule
\end{tabular}
\end{table}

\textbf{Impact of Temporal Memory Hubs.} Ablating the long-term workspace hub reduces performance by 1.3\%, confirming its role in retaining scene layouts and past object states. Similarly, removing the short-term hub leads to a 0.9\% drop, impairing high-frequency motion tracking. Eliminating both hubs causes a substantial 2.1\% decline, demonstrating that standard single-frame conditioning is fundamentally insufficient for the long-horizon tasks in our suite.

\textbf{Efficacy of MI-Driven Reasoning.} Removing the MI contrastive objective degrades performance by 0.8\%, suggesting that without explicit constraints, latent tokens fail to capture meaningful intent. Completely eliminating the reasoning bottleneck results in a 1.4\% drop in instruction following. This confirms that a dedicated semantic alignment layer is critical for bridging high-level linguistic goals with low-level control, preventing overfitting to superficial visual cues.

\subsection{Analysis of Temporal Grounding}
A core premise of MIRTH is that explicit memory hubs enable the model to overcome the temporal myopia of single-frame baselines (e.g., OpenVLA). To validate this claim and answer RQ3, we conduct two targeted analyses: probing the frozen representations for dynamic information and testing the model's sensitivity to temporal order. The results are shown in Table~\ref{tab:temporal_analysis}.

\begin{table}[t]
\centering
\small
\caption{Analysis of temporal grounding on LIBERO-Long simulation benchmark. We report mean absolute error (MAE) to measure motion awareness and success rate under temporal shuffling for history dependence.}
\label{tab:temporal_analysis}
\begin{tabular}{lcc}
\toprule
\textbf{Analysis} & \textbf{OpenVLA} & \textbf{MIRTH} \\
\midrule
\multicolumn{3}{l}{\textit{Linear Probing ($MAE$ of normalized proprio)}} \\
\quad state estimation & 0.32 & \textbf{0.11} \\
\quad velocity estimation & 0.15 & \textbf{0.04} \\
\midrule
\multicolumn{3}{l}{\textit{Frame Shuffling (success rate)}} \\
\quad standard inference & 53.7 & \textbf{95.3\%} \\
\quad shuffled history & -$^\dagger$ & 88.0\% \\
\bottomrule
\multicolumn{3}{l}{\footnotesize $^\dagger$Not applicable to single-frame baselines.}
\end{tabular}
\end{table}

\textbf{Linear Probing of Motion Dynamics.} If the temporal memory hubs effectively capture dynamics, the resulting embeddings should be linearly diagnostic of the robot's physical state changes. After training, we freeze MIRTH and train a lightweight regressor on top of $\tilde{Z}_t$ to predict the arm's instantaneous states and velocity (obtained by the difference between two frames). We compare this against a linear probe trained on the single-frame visual tokens of OpenVLA. The results indicates that while single-frame models capture static semantic layout (the estimated states are much closer), they fail to explicitly encode high-order derivatives of motion (the estimated velocity are much more offset). In contrast, MIRTH's workspace and short-term hubs successfully compress these dynamics into the latent space, providing the policy with the necessary velocity awareness for smooth control.

\textbf{Sensitivity to Frame Order.} To verify that MIRTH relies on the causal structure of history rather than treating memory as a bag of frames, we introduce a temporal shuffle during inference. Specifically, within the short-term memory hub, we randomly permute the order of the recent history frames while keeping the visual content identical. We observe that this perturbation causes a sharp decline in success rates, dropping by 7.3\% on LIBERO-long simulation. This sensitivity confirms that the model is not merely matching static visual textures from the past, but is actively exploiting the temporal coherence and sequential trends to infer future actions. The workspace hub, utilizing EMA, remains robust to high-frequency jitter but provides the necessary long-term context, further supporting our design of dual-scale temporal processing.

\subsection{Analysis of Latent Reasoning} 
We posit that the proposed MI objective encourages the latent reasoning tokens to capture high-level task semantics, serving as a compact bridge between instruction and control. To verify this and answer RQ4, we analyze the latent topology and the model's emergent behavior in failure scenarios.

\begin{figure}
    \centering
    \begin{minipage}{1.0\linewidth}
        \centering
        \includegraphics[width=1.0\textwidth]{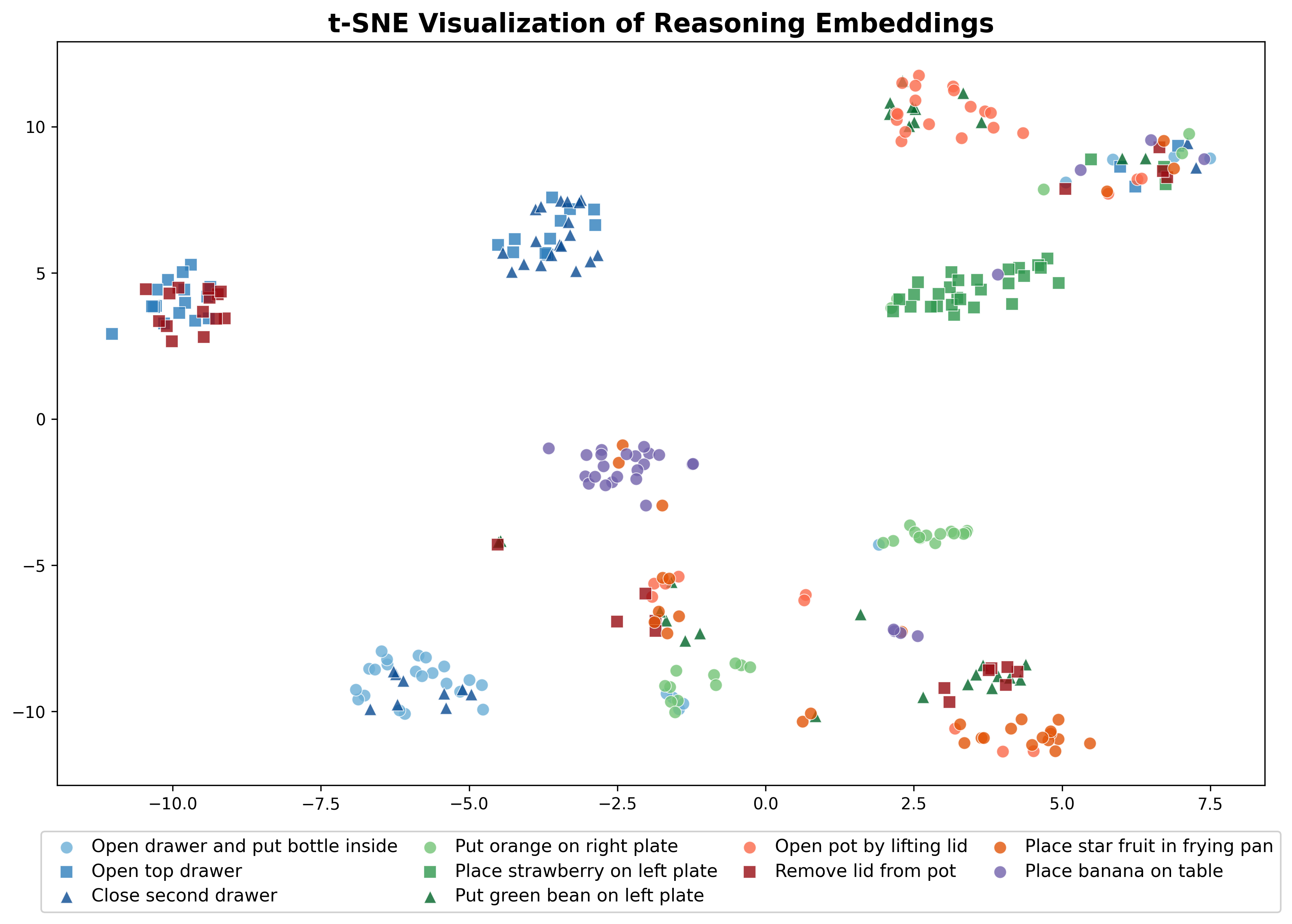}
    \end{minipage}
  \caption{\textbf{The t-SNE visualization of reasoning embeddings}. We select 10 tasks on LeRobot and run each task with 30 episodes.}
  \label{fig:tsne_visualization}
\end{figure}

\textbf{Semantic Organization of Reasoning Tokens.} To visualize the internal representations learned by MIRTH, we extract the reasoning token embeddings across 20 distinct tasks from the validation set and project them into a 2D space using t-SNE \citep{maaten2008visualizing}. As illustrated in Figure~\ref{fig:tsne_visualization}, even without explicit task-ID supervision, the reasoning tokens spontaneously organize into distinct clusters based on task semantics. For instance, tasks involving \textit{opening containers} form a coherent cluster, clearly separated from \textit{pick-and-place} tasks. This structural separation confirms that the MI objective successfully carves out a semantic plan space, where the model learns to group functionally similar behaviors together.

\textbf{Emergent Error Recovery and Re-planning.} The value of the structured reasoning space is most evident when the agent faces execution failures. To quantify this, we isolated evaluation episodes where the robot failed in some actions and measured the recovery rate. We present the comparative results in Table~\ref{tab:recovery_rates}. Standard single-frame baselines and the ablated models exhibit negligible recovery capabilities ($<10\%$). In contrast, MIRTH achieves a recovery rate of 12.1\%. This quantitative gap confirms that the reasoning tokens effectively function as a dynamic discrepancy checker. When the tactile or visual feedback conflicts with the internal plan, the reasoning module halts the open-loop sequence and triggers a re-grasping maneuver, effectively converting potential failures into successes through closed-loop control.

\begin{table}[t]
\centering
\small
\caption{\textbf{Error Recovery Comparison.} We report the recovery rate on LeRobot failure scenarios. A successful recovery is defined as detecting the failure, re-attempting the action, and completing the task within the time budget.}
\label{tab:recovery_rates}
\begin{tabular}{lc}
\toprule
\textbf{Method} & \textbf{Recovery Rate} \\
\midrule
OpenVLA (single-frame) & 5.2\% \\
MIRTH (w/o reasoning tokens) & 8.7\% \\
\textbf{MIRTH (full model)} & \textbf{12.1\%}\\
\bottomrule
\end{tabular}
\end{table}

\section{Conclusion}
In this work, we presented MIRTH, a unified VLA framework designed to overcome several limitations of current architectures. By utilizing dual-scale temporal memory hubs, MI-guided latent reasoning, and parallel action decoding scheme, MIRTH effectively addresses the challenges of temporal myopia, reasoning gaps, and inference throughput. Our extensive evaluations on the LIBERO simulation benchmark and real-world LeRobot platforms demonstrate that MIRTH not only achieves SOTA success rates but also enables high-frequency control. Looking forward, we plan to extend this framework by incorporating multi-sensory modalities, such as tactile feedback, to further enhance manipulation precision in unstructured and occluded settings.

\section*{Limitations}
While MIRTH demonstrates promising capabilities, several limitations remain. 
First, regarding interpretability, unlike methods that utilize textual Chain-of-Thought, our latent reasoning tokens are optimized for information flow rather than human readability. While they effectively bridge perception and action, their opaque nature complicates the debugging process and makes it challenging to explicitly audit the model's internal decision logic prior to execution.
Second, regarding embodiment scope, our current experimental validation is restricted to stationary single-arm manipulation. Extending MIRTH to bimanual systems or mobile manipulators introduces additional complexities in coordination and whole-body control that were not addressed in this study.
Finally, although we utilize temporal memory hubs, the model's long-horizon reasoning capability is still bounded by the fixed size of the compressed memory prompts. Extremely long tasks requiring the retention of state information from hundreds of steps ago may still suffer from catastrophic forgetting.

\section*{Ethical Considerations}
As an embodied AI system capable of physical interaction, MIRTH presents potential safety risks if deployed without adequate safeguards. While our parallel decoding scheme improves real-time response, the model may still exhibit unpredictable behaviors in out-of-distribution scenarios. We strongly advise that any real-world deployment be accompanied by hardware-level emergency stop mechanisms and human supervision. MIRTH leverages pre-trained VLMs as backbones. Consequently, it may inherit social biases present in the web-scale pre-training data. Although our fine-tuning focuses on robotic manipulation, there is a residual risk that the model might exhibit biased behaviors when interpreting instructions related to culturally sensitive objects or scenarios. Our real-world data collection and evaluation adhered to strict privacy protocols. No personally identifiable information or human faces were explicitly targeted or retained in the training datasets.

\section*{Acknowledgments}
This work was supported by JST CREST, Japan, under Grant JPMJCR25T4. Shiyu Teng would like to thank the Program for Forming Japan's Peak Research Universities (J-PEAKS) (Grant No. R6-20) for supporting his postdoctoral position. 
The first author would like to acknowledge the complete freedom and independent environment provided during this research.

\bibliography{custom}

\clearpage
\appendix

\begin{figure*}[ht]
    \centering
    \begin{minipage}{1.0\linewidth}
        \centering
        \includegraphics[width=1.0\textwidth]{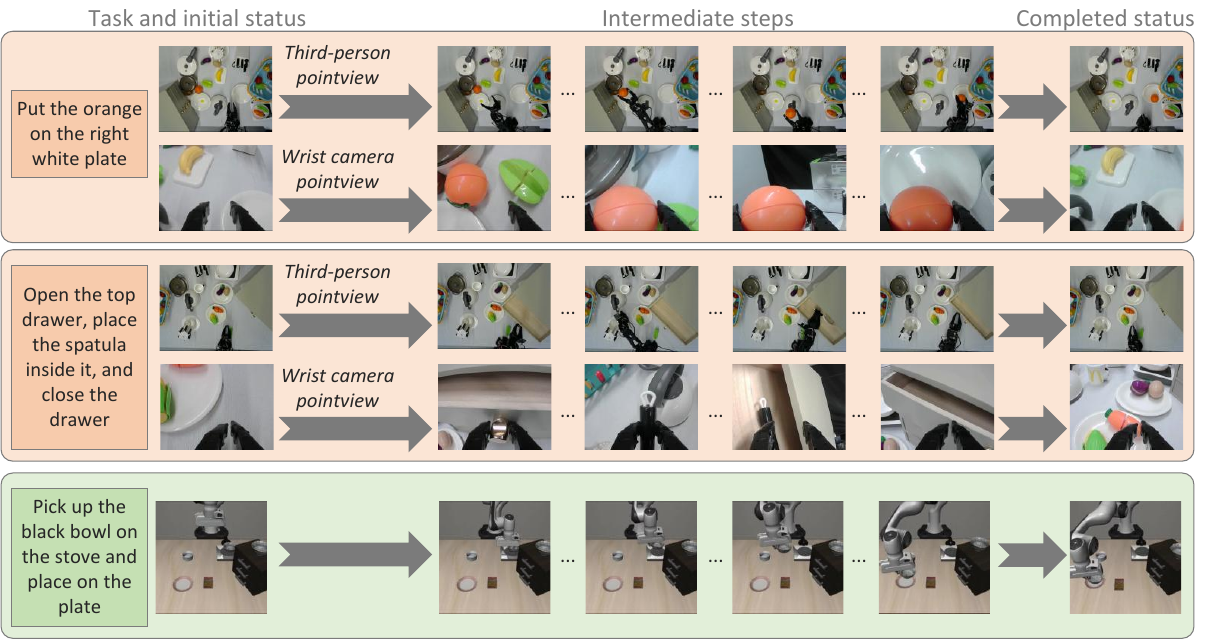}
    \end{minipage}
  \caption{Some visualized rollouts of MIRTH. We showcase successful execution trajectories on both physical and simulated platforms. \textbf{Top Rows (LeRobot real-world):} Two distinct tasks are shown with synchronized third-person and wrist-camera views. The first row depicts a spatial precision task, while the second row demonstrates a long-horizon reasoning task involving articulated objects. \textbf{Bottom Row (LIBERO simulation):} A sample rollout from the simulation benchmark, confirming MIRTH's sim-to-real consistency.}
  \label{fig:rollout_visualization}
\end{figure*}

\section{Experimental Settings}~\label{sec:setup}
All models were trained using mixed-precision on a compute node equipped with two NVIDIA RTX Pro 6000 GPUs. We used a global batch size of 64 (each GPU with 32) and it takes about five days to tune a single model. For the workspace memory hub, we employ $K=4$ distinct temporal scales. The decay rates $\{\beta_k\}_{k=1}^K$ are logarithmically spaced between $\beta_{\min}=0.01$ and $\beta_{\max}=0.3$, formally defined as:
\begin{equation}
\beta_k = \exp\left(\ln \beta_{\min} + \frac{k-1}{K-1}(\ln \beta_{\max} - \ln \beta_{\min})\right)
\end{equation}
yielding $\beta_k \approx \{0.01, 0.0311, 0.0965, 0.3\}$. The update coefficients for velocity and variability (Eq.~2) are set to $\gamma_\mu=0.2$ and $\lambda_\sigma=0.2$, respectively. For the short-horizon memory hub, we set the window size $w=4$, corresponding to a temporal receptive field of approximately 0.4 seconds at a control frequency of 10Hz. In the attention mechanism (Eq.~7), we use a temperature scaling of $\tau_r=1.0$ and a recency bias factor of $1.1$. To balance the auxiliary reasoning objectives with the primary behavior cloning loss, we set the mutual-information coefficients $\lambda_{ra}$ and $\lambda_{rx}$ (Eq.~15) to $1.0$. The contrastive regularization term $\lambda_{mi}$ (Eq.~17) is scaled to $0.001$ to stabilize the learning of the reasoning latent space without dominating the gradient updates.

During training, the pretrained VLM backbone remains largely frozen. We apply Low-Rank Adaptation (LoRA) to the LLM backbone with rank $32$ to efficiently fine-tune the multimodal representations for embodied control without dropout. In total, the MIRTH architecture comprises approximately 8.02 billion parameters, of which only 482.34 million (approximately 6.01\%) are trainable. This parameter-efficient fine-tuning strategy is distributed as follows:
\begin{itemize}
    \item OpenVLA Backbone Components: The core LLM backbone is fine-tuned via LoRA, updating 79.95M parameters (16.58\% of the trainable total). The heavy vision backbone (730.91M parameters) and the main visual projector remain entirely frozen to preserve pretrained semantic knowledge. The proprioceptive projector is fully trainable, accounting for 16.82M parameters.
    
    \item Action Decoder: The lightweight projection head used for parallel action decoding introduces 285.74M trainable parameters. Because it replaces the autoregressive generation pipeline, it accounts for the largest portion (59.24\%) of the tunable parameter pool.

    \item Temporal Memory Hubs: The modules responsible for predicting per-patch mixture weights and projecting temporal statistics for the dual-scale memory hubs introduce 46.86M trainable parameters.

    \item Latent Reasoning Tokens: The components dedicated to the mutual-information alignment and latent reasoning tokens contribute an additional 52.96M trainable parameters.
\end{itemize}

\section{Experimental Setup}
\textbf{LIBERO Simulation Benchmark.} We utilize the LIBERO benchmark suite \citep{liu2023libero}, which evaluates agents on long-horizon robustness and generalization. The benchmark consists of four distinct suites: LIBERO-Spatial, LIBERO-Object, LIBERO-Goal, and LIBERO-Long, with each suite containing 10 specific tasks.For data preprocessing, we filter out static idle frames and resize all visual observations to $224 \times 224$. To align with the pretraining distribution of our VLA backbone, we apply a vertical flip augmentation to input images. During evaluation, we conduct 30 rollout trials for each task with randomized initializations and report the average success rate.

\textbf{LeRobot Real-Robot Evaluation.} We deploy MIRTH on a physical single-arm manipulator using the open-source LeRobot platform (detailed hardware specifications are provided in Appendix A). We curated a diverse dataset covering 126 distinct tasks, ranging from basic pick-and-place primitives to complex multi-stage reasoning scenarios (in Appendix). For each task, we collected 3 expert demonstrations, ensuring robustness by randomizing initial object poses and environment configurations between episodes. To improve optimization stability and training efficiency, we adopted a task clustering strategy where semantically similar tasks are grouped and trained jointly. Consistent with our simulation setup, we perform 30 evaluation trials for each target task in the real world to calculate the final success rate. The evaluations are conducted on one RTX 5090 GPU.

\begin{figure*}[ht]
    \centering
    \begin{minipage}{1.0\linewidth}
        \centering
        \includegraphics[width=1.0\textwidth]{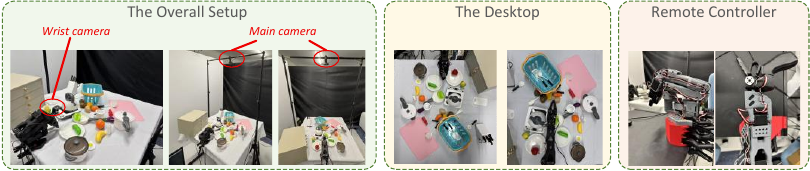}
    \end{minipage}
  \caption{The LeRobot setup, including the whole environment, the desktop and the second remote controller. }
  \label{fig:lerobot_setup}
\end{figure*}

\section{LeRobot Setup}
Figure~\ref{fig:lerobot_setup} illustrates our physical experimental environment utilizing the LeRobot platform. The setup features a robotic manipulator with an operating radius of approximately 55 cm. To simulate diverse real-world manipulation scenarios, we arrange a variety of kitchen-themed objects on a $100 \times 100$ cm white tabletop. To rigorously evaluate MIRTH's reasoning capabilities involving articulated objects and containment, the workspace is further augmented with a multi-drawer cabinet and a bucket (shown in the center of Figure~\ref{fig:lerobot_setup}). The perceptual system comprises two RGB cameras: a \textit{wrist-mounted camera} for fine-grained, egocentric observation, and a \textit{main overhead camera} positioned approximately 115 cm above the table center to capture global context (shown in the left of Figure~\ref{fig:lerobot_setup}). For collecting expert demonstrations, we employ a high-fidelity teleoperation framework where a secondary leader arm (shown in the right panel of Figure~\ref{fig:lerobot_setup}) controls the operating robot in real-time.

\section{Some Visualization}
To intuitively demonstrate the efficacy of MIRTH, we visualize representative policy rollouts across both the LeRobot real-world platform and the LIBERO simulation benchmark in Figure~\ref{fig:rollout_visualization}. The visualizations track the agent's progress from the initial state through critical intermediate manipulation steps to successful task completion. As illustrated, MIRTH demonstrates stable control not only in atomic pick-and-place operations but also in complex, multi-stage reasoning tasks (e.g., manipulating articulated objects).

\begin{table}[ht] 
\centering 
\small
\caption{Comparison of memory integration strategies on LIBERO-long, including both performance and efficiency.} 
\label{tab:prefix_infusion} 
\begin{tabular}{lcc} 
\toprule 
\textbf{Method} & \textbf{Success Rate} & \textbf{Throughput} \\
\midrule 
Prefix & \textbf{95.3\%} & 64.4 Hz \\
Infusion & 93.1\% & \textbf{70.0 Hz} \\ 
\bottomrule 
\end{tabular} 
\end{table}

\section{Memory Integration Strategies (Prefix vs. Infusion)}\label{sec:prefixvsinfusion}
In the main architecture of MIRTH, the compressed history features from the temporal hubs must be injected into the VLM backbone. We investigated two distinct architectural strategies for this integration:

\begin{itemize}
    \item Prefixing: The fused memory features are projected and appended to the input sequence prior to the language instruction. This allows the full self-attention mechanism of the VLM to attend to historical details but increases the effective sequence length. 
    \item Infusion: The memory features are injected directly into the current frame's visual patch embeddings via a lightweight linear infusion. As shown in Eq.11, this method maintains a constant sequence length regardless of memory capacity. 
\end{itemize}

We compare these two strategies on LIBERO-Long benchmark in Table~\ref{tab:prefix_infusion}. As indicated by the results, the prefixing strategy achieves a higher success rate (2.2\%). We argue that allowing the LLM to explicitly attend to memory tokens as distinct entities facilitates more robust temporal reasoning, particularly for long-horizon recall. However, the infusion offers a significant advantage in computational efficiency, achieving a higher control frequency by avoiding the quadratic cost associated with longer context windows. For our main experiments, we prioritized performance and utilized the prefix manner, while infusion remains a viable alternative for resource-constrained deployment scenarios.

\section{Ablation on Decoding Paradigms}\label{sec:action_decoding}
(This section is marked with a separate symbol from the other parts for better understanding.)

We also investigated the optimal strategy for mapping the VLM's latent representations to continuous control signals. Specifically, given a target action chunk of length $T$ with $F$ degrees of freedom (DoF), we define the ground-truth action trajectory as $\mathbf{A} \in \mathbb{R}^{T \times F}$. Let $\mathbf{H} \in \mathbb{R}^{N \times D}$ denote the output hidden states from the language model dedicated to action prediction, where $N$ is the number of action tokens and $D$ is the embedding dimension. We compare four distinct decoding paradigms:

\begin{enumerate}

\item \textbf{Scalar-wise decoding (standard VLA):} Each token represents a single scalar action dimension.
    \begin{itemize}
        \item $N = T \times F$. (e.g., for $T=10, F=6$, we require 60 tokens in context modeling).
        \item The model autoregressively predicts discretized bins or scalar values for each degree of freedom sequentially.
    \end{itemize}
    
\item \textbf{Global vector-wise decoding (concatenated):} Each token represents a single timestep, and the entire sequence is projected jointly.
    \begin{itemize}
        \item $N = T$.
        \item We flatten the hidden states of all timesteps into a single vector and apply a global projection matrix $\mathbf{W}_{\text{global}} \in \mathbb{R}^{(T \cdot D) \times (T \cdot F)}$:
        \begin{equation}
            \hat{\mathbf{A}}_{\text{flat}} = \mathbf{W}_{\text{global}} \cdot \text{flatten}(\mathbf{H})
        \end{equation}
        where $\text{flatten}(\mathbf{H}) \in \mathbb{R}^{T \cdot D}$ is the flattened representation. This allows the decoding head to capture inter-timestep dependencies explicitly.
    \end{itemize}

\item \textbf{Independent vector-wise decoding (parallel):} Each token represents a single timestep, but is projected independently.
    \begin{itemize}
        \item $N = T$.
        \item A shared projection matrix $\mathbf{W}_{\text{sep}} \in \mathbb{R}^{D \times F}$ is applied to each token's hidden state $\mathbf{h}_t$ in parallel:
        \begin{equation}
            \hat{\mathbf{a}}_t = \mathbf{W}_{\text{sep}} \cdot \mathbf{h}_t, \quad t \in \{1, \dots, T\}.
        \end{equation}
    \end{itemize}

\item \textbf{Condensed chunk decoding:} A single token represents the entire trajectory chunk.
    \begin{itemize}
        \item $N = 1$.
        \item The single hidden state $\mathbf{H} \in \mathbb{R}^{D}$ is projected to the full trajectory via $\mathbf{W}_{\text{chunk}} \in \mathbb{R}^{D \times (T \cdot F)}$:
        \begin{equation}
            \hat{\mathbf{A}}_{\text{flat}} = \mathbf{W}_{\text{chunk}} \cdot \mathbf{H}.
        \end{equation}
    \end{itemize}
    
\end{enumerate}

\begin{figure*}[ht]
    \centering
    \begin{minipage}{1.0\linewidth}
        \centering
        \includegraphics[width=0.8\textwidth]{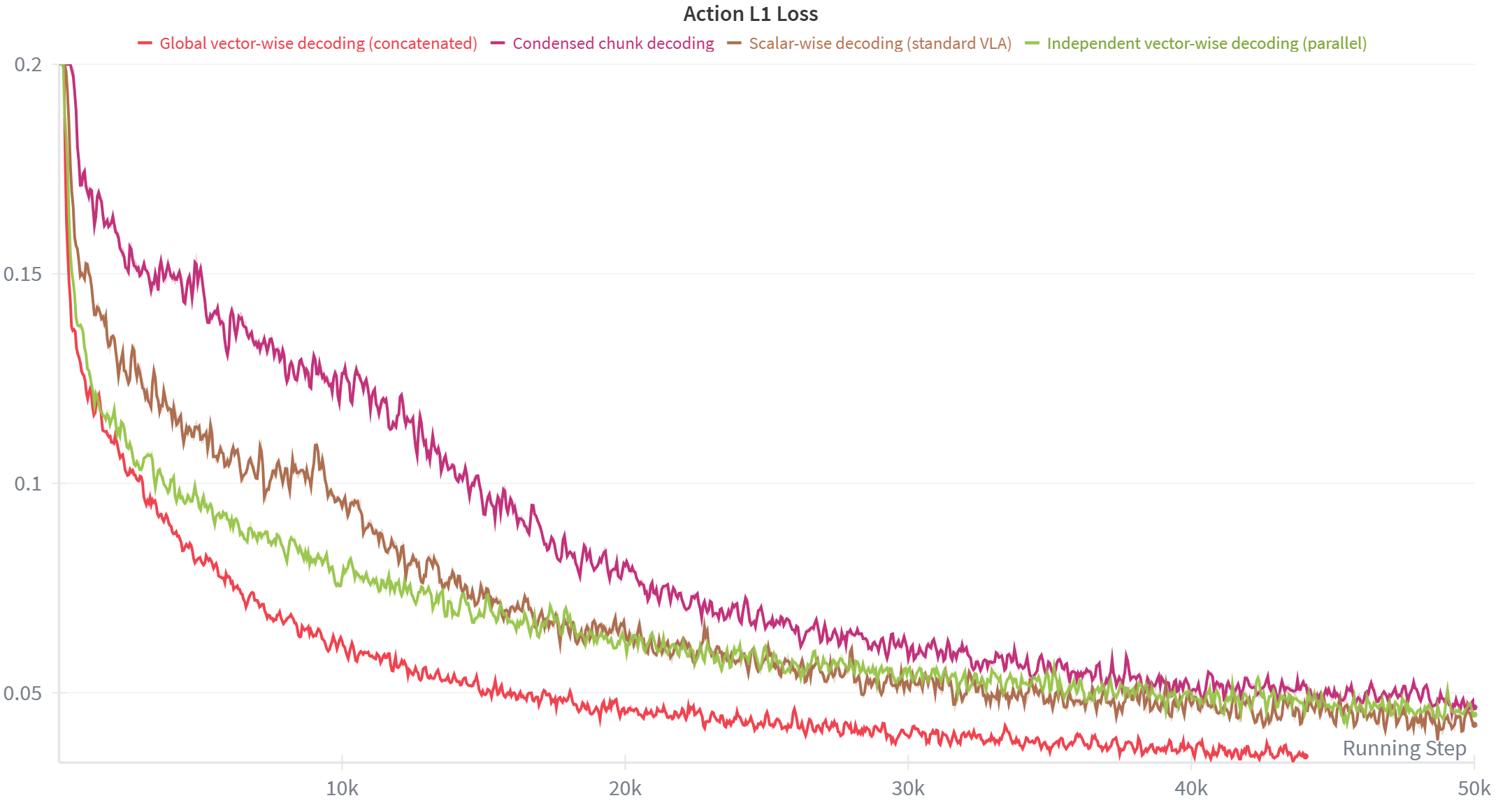}
    \end{minipage}
  \caption{The comparision of convergence for four different decoding paradigms on LIBERO-object.}
  \label{fig:decoding_converge}
\end{figure*}

\begin{figure*}[ht]
    \centering
    \begin{minipage}{1.0\linewidth}
        \centering
        \includegraphics[width=1.0\textwidth]{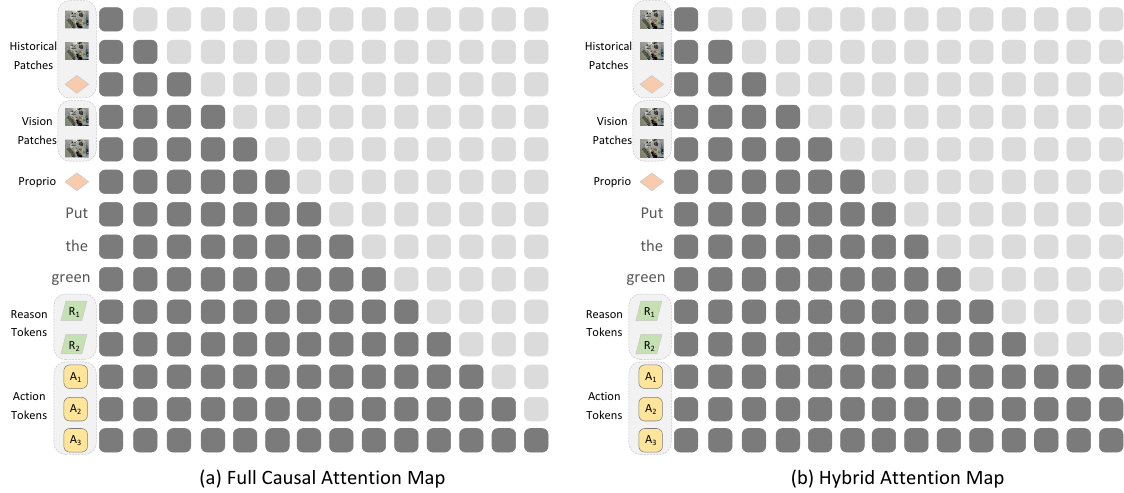}
    \end{minipage}
  \caption{The illustration of full causal attention map and hybrid attention map.}
  \label{fig:attention_map}
\end{figure*}

In our experiments, we observe significant differences in convergence dynamics among these paradigms (but with comparable final performance). As visualized in Figure~\ref{fig:decoding_converge}, the second method (global concatenated vector-wise decoding) exhibits the fastest convergence rate. We hypothesize that this is because it balances representation capacity with decoding complexity. Conversely, the fourth method is the slowest to converge due to this compression burden. Consequently, we adopt the global concatenated vector-wise decoding strategy in our final MIRTH architecture.

\section{Effect of Action Chunk Size}\label{sec:chunk_ablation}
We analyze the impact of the action chunk size (the number of consecutive timesteps predicted in a single forward pass) on both task performance and system efficiency. This hyperparameter introduces a critical trade-off between control precision and computational speed. As detailed in Table~\ref{tab:chunk_ablation}, we evaluate the model with chunk sizes ranging from 1 to 30. We observe two distinct trends: (1) Throughput increases linearly with chunk size. Since the computational cost is dominated by the heavy vision encoder and VLM backbone, generating a chunk of 30 actions incurs roughly the same cost as generating a single action. Larger chunks effectively amortize this backbone computation over more timesteps. (2) Performance peaks at smaller chunk sizes but degrades at extremes. When setting chunk size to 1, the system suffers from low temporal coherence. However, at chunk size 30, the success rate drops due to the optimization difficulty. Based on this analysis, we select chunk size to 10 as the optimal equilibrium, which is also the sampling rate of our collected trajectories (10 Hz).

\begin{table}[ht]
\centering
\small
\caption{\textbf{Ablation on Action Chunk Size.} We report the success rate on the LeRobot platform (task: \textit{place the green bean to the right white plate}) and the inference throughput on a single RTX 5090 GPU. Larger chunks improve speed via amortization but degrade performance.}
\label{tab:chunk_ablation}
\begin{tabular}{ccc}
\toprule
\textbf{Chunk Size} & \textbf{Success Rate} & \textbf{Throughput} \\
\midrule
1 & 94.0 & 41 Hz \\
\textbf{5} & \textbf{95.0} & 58 Hz \\
10 (ours) & 94.0 & 62 Hz \\
20 & 92.0 & 65 Hz \\
30 & 88.0 & \textbf{68 Hz }\\
\bottomrule
\end{tabular}
\end{table}

\section{Full Causal Attention vs Hybrid Attention}
We analyze the impact of the attention masking strategy on both training efficiency and model performance. In VLA architectures, two common masking schemes are available: (1) Hybrid attention, where action tokens are allowed to attend bidirectionally to each other. This theoretically allows for richer context modeling within the prompt. (2) Full causal attention, which is a standard lower-triangular mask is applied to the entire sequence, enforcing a strict causal dependency from left to right for all tokens. The differences are shown in Figure~\ref{fig:attention_map}.

While hybrid attention intuitively offers better context modeling, it introduces significant engineering bottlenecks. Most SOTA opensource optimization kernels, such as FlashAttention-2, are highly optimized for standard causal masking but do not natively support arbitrary custom masks without degrading to slower kernels. In our experiments, implementing hybrid attention required bypassing FlashAttention and relying on standard PyTorch SDPA (Scaled Dot-Product Attention) with explicit mask tensors. We observed that this configuration resulted in significantly slower training convergence and reduced computational throughput compared to the optimized causal kernel.

In contrast, full causal attention is fully compatible with standard FlashAttention acceleration. Empirically, we found that the strict causal constraint does not hamper the model's ability to reason about the visual context, as the model learns to aggregate necessary information into later tokens. Given the negligible performance difference and the substantial gain in training speed, we adopt full causal attention as the default configuration for MIRTH.

\section{LeRobot Dataset and Evaluation Protocols}\label{sec:tasks}
To facilitate reproducibility and rigorous evaluation, we detail the data collection process and the task definitions used in our real-world LeRobot experiments.

\subsection{Collected Training Dataset}
We collected a total of 1000 expert trajectories via teleoperation on the LeRobot platform. The dataset is structured into five distinct task groups of increasing complexity, ranging from atomic manipulation to semantic reasoning. For each task, we collected 50 demonstrations with randomized initial object poses. The detailed task list is provided in Table~\ref{tab:training_tasks}.

\begin{table*}[ht]
\centering 
\small 
\caption{\textbf{LeRobot Training Dataset Details.} We collected 50 trajectories for each of the 20 tasks, totaling 1000 episodes. The tasks are categorized into five groups based on the required manipulation skills and reasoning complexity.} 
\label{tab:training_tasks} 
\begin{tabular}{lp{0.6\textwidth}c} 
\toprule 
\textbf{Task Group} & \textbf{Instruction (Training Set)} & \textbf{Count} \\
\midrule
\multirow{4}{*}{\textbf{1. Basic Tasks}} & Place the banana in the plate on the right & 50 \\
& Place the brown kiwi on the cutting board & 50 \\
& Place the carrot in the plate on the left & 50 \\
& Place the star fruit in the white frying pan & 50 \\
\midrule 
\multirow{4}{*}{\textbf{2. Mechanism Operations}} & Open the top drawer of the four-drawer cabinet & 50 \\
& Close the second drawer of the four-drawer cabinet & 50 \\
& Open the top drawer, place the spatula inside it, and close the drawer & 50 \\
& Open the second drawer, put the banana into it, and close the drawer & 50 \\
\midrule 
\multirow{4}{*}{\textbf{3. Scene Rearrange}} & Empty the small bucket onto the cutting board & 50 \\
& Swap all items currently on the left white plate with the items on the right white plate & 50 \\
& Clear the white frying pan by moving any items inside it onto the cutting board, leaving the frying pan empty & 50 \\
& Clean up the workspace by moving all fruits onto the left white plate and all vegetables onto the right white plate & 50 \\
\midrule 
\multirow{4}{*}{\textbf{4. Category Reasoning}} & Put all fruits except the banana into the white frying pan & 50 \\
& Clear the cooking area: move all food items off the cutting board and leave only tools on the cutting board & 50 \\
& Place all vegetables except the corn with green leaves into the pot with the dark lid & 50 \\
& Move any fruits that are directly on the table into the pot with the dark lid and any vegetables that are directly on the table into the white frying pan & 50 \\
\midrule 
\multirow{4}{*}{\textbf{5. Semantic Recipe}} & Prepare ingredients for a simple vegetable scramble by placing the raw egg, carrot, green bean, and yellow bell pepper onto the cutting board, and leave all fruits where they are & 50 \\
& Prepare ingredients for a fruit yogurt by placing the strawberry, kiwi, apple pieces, and banana into the white frying pan & 50 \\
& Prepare a 'breakfast plate' by placing the cooked fried egg, one fruit, and one vegetable together on the right white plate & 50 \\
& Put all fruits that are good for a refreshing snack, the orange, kiwi, strawberry, and star fruit, on the left white plate & 50 \\ 
\bottomrule 
\end{tabular} 
\end{table*}

\subsection{Evaluation on Unseen Instructions} To assess MIRTH's generalization capability, we also designed a set of validation tasks. These tasks share the same semantic logic as the training groups but involve unseen object combinations, different target locations, or paraphrased linguistic goals. This setup ensures the model is not merely overfitting to specific sentence patterns or memorized trajectories. The generated validation instructions are listed in Table~\ref{tab:validation_tasks}.

\begin{table*}[ht] 
\centering 
\small 
\caption{\textbf{Validation Tasks on LeRobot.} For each training group, we evaluate the model on semantically similar but distinct instructions to test generalization.} 
\label{tab:validation_tasks} 
\begin{tabular}{lp{0.7\textwidth}} 
\toprule 
\textbf{Group} & \textbf{Validation Instructions} \\
\midrule \textbf{1. Basic Tasks} & Put the carrot on the cutting board. \\
\midrule 
\textbf{2. Mechanism Ops} & Open the second drawer, place the soup spoon inside, and then close it. \\
\midrule 
\textbf{3. Scene Rearrange} & 
Organize the tools by placing the spatula, spoon, and strainer together in front of the cabinet. \\
\midrule 
\textbf{4. Category Reasoning} & Place all red items (apple, strawberry, red bottle) onto the left white plate and all green items onto the cutting board. \\
\midrule 
\textbf{5. Semantic Recipe} & Prepare a 'healthy dinner' set by placing the corn, carrot, and eggplant into the pot with the dark lid, while keeping the fruits on the table. \\
\bottomrule 
\end{tabular} 
\end{table*}

\end{document}